\documentclass{svproc}
\usepackage{microtype}
\usepackage{adjustbox}
\usepackage{amsmath}
\usepackage{graphicx}
\usepackage{indentfirst}
\usepackage{hyperref}

\begin{document}
\title{Traffic Sign Classification Using Deep and Quantum Neural Networks}
%
%
\author{Sylwia Kuros \and Tomasz Kryjak \href{https://orcid.org/0000-0001-6798-4444}{\includegraphics[width=16pt]{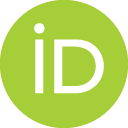}}}
\authorrunning{S. Kuros, T. Kryjak} 
%
%
\institute{Embedded Vision Systems Group, Computer Vision Laboratory, \\ Department of Automatic Control and Robotics, \\ AGH University of Science and Technology, Kraków, Poland\\
\email{sylwiakuros@student.agh.edu.pl, tomasz.kryjak@agh.edu.pl}
}
\maketitle              
\begin{abstract}
Quantum Neural Networks (QNNs) are an emerging technology that can be used in many applications including computer vision. In this paper, we presented a traffic sign classification system implemented using a hybrid quantum-classical convolutional neural network. Experiments on the German Traffic Sign Recognition Benchmark dataset indicate that currently QNN do not outperform classical DCNN (Deep Convolutuional Neural Networks), yet still provide an accuracy of over 90\% and are a definitely promising solution for advanced computer vision. 

\begin{center}
\keywords{Quantum Neural Networks, Traffic Sign Recognition, DCNN, GTSRB}
\end{center}
\end{abstract}
%
%
%
\section{Introduction}
\label{introduction}

\begin{sloppypar}
\setlength\parindent{24pt}

Nowadays, the amount of data produced doubles every two years \cite{gallagher}, and the~peak of the heyday of computers in the~classical meaning is coming to an end. 
Maintaining the current momentum of technological development requires a~change in the approach to computing. 
One of the most promising solutions is to transfer the idea of computing from the field of classical mechanics to quantum mechanics, which creates new and interesting possibilities, but at the same time poses significant challenges.

In computer vision systems that for effective operation require processing of large amounts of data in real time, quantum neural networks (QNNs) can prove to be a very attractive solution.
On the basis of experiments carried out in recent years, it can be observed that the network training process is becoming quicker and results are more accurate. 
They also show better generalization capabilities even with small amounts of training data and require several times fewer epochs compared to classical networks.
QNNs were successfully used, among others, in the following applications: tree recognition in aerial space of California \cite{boyda}, cancer recognition \cite{li}, facial expression recognition \cite{li2}, vehicle classification~\cite{yu}, traffic sign recognition from the LISA database considering the vulnerability of adversarial attacks \cite{majumder}, handwriting recognition \cite{zhou}, \cite{oh}, \cite{potempa}, \cite{zhao}, \cite{henderson}, \cite{hernandez}, object segmentation \cite{aytekin}, \cite{aytekin2},  pneumonia recognition \cite{yumin}, classification of ants and bees \cite{mari}, classification of medical images of chest radiography and retinal color of the fundus \cite{mathur}, and generative networks~\cite{lloyd},~\cite{dallaire}.

In this paper, we describe the results of our work on a quantum neural network for the classification of traffic signs from the German Traffic Sign Recognition Benchmark (GTSRB) dataset.
The aim of our research was to analyze and compare the results obtained using a classical deep convolutional neural network (DCNN) and a quantum neural network.  
The architecture was implemented using the Python language and the PennyLane library for quantum computing. 
To the best of our knowledge, this is the only work on traffic sign classification for the popular GTSRB dataset that uses quantum computing techniques. 
Another paper (\cite{potempa}) on a similar topic deals with a~different set and the proposed methods have lower efficiency.

The remainder of this paper is organized as follows. 
Section \ref{qnn} provides background information on quantum neural networks. 
In Section \ref{previouswork} our motivation and the purpose of our experiment were described in the context of related work on machine learning applied to computer vision systems. 
Section \ref{qnnfortsr} presents the experiments conducted and the results obtained. 
In Section \ref{conclusion} conclusions and ideas for further development are provided.

\section{Quantum Neural Networks}
\label{qnn}

The registers of quantum computers can exist in all possible states simultaneously, due to the property of superposition, with a chance of capturing a state at the time of measurement whose probability was encoded before the measurement. 
Quantum computing works on the principle of increasing the probability of a~desired state to a certain high value, so that this state can be reached with high confidence and with as few measurements as possible. \par

In this context, quantum interference resulting from the superposition phenomenon allows the amplitudes of the probabilities corresponding to given states to influence each other. The same is true for quantum entanglement, which allows quantum objects to be linked by a strong correlation, useful for readings of quantum states.
Quantum entanglement means that two particles remain connected despite the distances that separate them. 
In quantum computing, such a state allows for the connection of a large number of qubits (quantum equivalent of the classical bit) together, thereby increasing computational resources in a~nonlinear way. \par

There are many approaches to the issue of physical prototyping of a quantum computer, the two most popular being based on superconducting electronic circuits or ion traps. 
In the case of superconducting quantum circuits, the quantum processor is placed at the~bottom of the gas pedal cylinder in a shield composed of a material called cryoperm and a magnetic field. It is led by wires that send microwave pulses of varying frequency and duration to control and measure the state of the qubits. Due to the interference that temperature and environmental factors bring, dilution chillers are used to lower the temperature to 15 millikelvin. Quantum information can also be destroyed by resistance, so a superconducting material with zero resistance at some low temperature is used~\cite{huang}. \par

A slightly different operating principle can be observed in quantum computer architectures based on ion traps. 
The advantage of this approach is the ability to capture the state of a qubit at a well-defined location \emph{trapping} it using an electromagnetic trap called a Paul trap \cite{paul}. 
It works on the principle of a~time-varying electric field in such a way as to hold the ion in~constant position. 
To obtain architectures with multiple qubits, it is necessary to trap multiple ions in a linear chain. 
The use of ions as qubits requires the use of laser light to change the state of an electron in an atom from a ground state to an excited state. 
Keeping it in the~excited state is achieved by changes in the laser frequency. \par

Due to the lack of widespread access to real quantum computers, a lot of research is carried out in an artificial environment using special software that simulates the ideal relationships found in quantum mechanics. 
In the case of this paper, considerations based on a simulated environment have also been performed.

Currently available classical computers, including supercomputers and cloud computing, allow simulating quantum hardware.
This supports, to a limited extent, research on quantum algorithms and computations in parallel to the work on the hardware layer.
One of the interesting topics are quantum neural networks addressed in this paper.

In its current stage of development, quantum deep learning is based on two classes of neural networks. 
The first one consists of quantum and classical neural layers and is called a hybrid quantum-classical neural network, whereas the second uses only quantum gates in layer construction. \par

A hybrid quantum-classical neural network contains a hidden quantum layer built of parameterized quantum circuits whose basic units are quantum gates. 
They allow the state of the qubit to be modified by superposition, entanglement, and interference, so that the measurement result uniquely correlates with the unobservable state. 
In quantum neural networks, quantum gates rotate the state of a qubit that is a parameter for rotating gates in a given layer, based on the output of a classical circuit in the preceding layer. 

The idea of training a classical and a quantum network is the same. However, an important difference is that, in the case of quantum networks, the parameters of the quantum circuit are optimized, whereas, in the process of learning a~classical network, the search for the best possible weights is performed. 



There are an infinite number of quantum gates, but this paper focuses on discussing the most relevant ones.
The quantum NOT gate (or Pauli-X) belongs to the class of single-qubit gates. 
For visualization purposes, the way to represent a qubit states in a 3D spherical coordinates called the Bloch sphere was proposed. 
The operation it performs corresponds to the rotation of the \(x\) on the Bloch sphere by an angle \(\pi\). 
In other words, the NOT gate assigns a probability amplitude to the state \(|0 \rangle\) to \(|1 \rangle\) and vice versa according to the following formula:
\begin{equation}
\label{eqn:paulix}
X : \alpha|0 \rangle + \beta|1 \rangle \rightarrow \beta|0 \rangle + \alpha|1 \rangle
\end{equation}

The matrix representation is as follows:
\begin{equation}
\label{eqn:diracstate_1}
X = \begin{bmatrix}
0 & 1\\
1 & 0
\end{bmatrix}
\end{equation}

The Pauli-Y gate corresponding to the rotation of the \(y\) axis on the Bloch sphere by an angle \(\pi\) is described by a unitary matrix of the following form:
\begin{equation}
\label{eqn:diracstate_2}
Y = \begin{bmatrix}
0 & -i\\
i & 0
\end{bmatrix}
\end{equation}

The Pauli-Z gate for the rotation of the \(z\) axis on the Bloch sphere by the angle \(\pi\) is described by the following unitary matrix:
\begin{equation}
\label{eqn:diracstate_3}
Z = \begin{bmatrix}
1 & 0\\
0 & -1
\end{bmatrix}
\end{equation}

Similarly, the rotation gates \(RX\),\(RY\),\(RZ\) belonging to the group of single-qubit gates perform an angle rotation \(\theta\) on the Bloch sphere, respectively, around the axis \(x\), \(y\), \(z\). 

In matrix representation \(RX\) is given as follows:
\begin{equation}
\label{eqn:ry_1}
RX(\theta) = \begin{bmatrix}
\cos\frac{\theta}{2} & -\sin\frac{\theta}{2}\\
\sin\frac{\theta}{2} & \cos\frac{\theta}{2}
\end{bmatrix}
\end{equation}

Gate \(RY\):
\begin{equation}
\label{eqn:ry_2}
RY(\theta) = \begin{bmatrix}
\cos\frac{\theta}{2} & -i\sin\frac{\theta}{2}\\
-i\sin\frac{\theta}{2} & \cos\frac{\theta}{2}
\end{bmatrix}
\end{equation}

Gate \(RZ\):
\begin{equation}
\label{eqn:ry_3}
RZ(\theta) = \begin{bmatrix}
e^{-i\frac{\theta}{2}} & 0\\
0 & e^{i\frac{\theta}{2}}
\end{bmatrix}
\end{equation}

The Hadamard gate is also a gate that operates on a single qubit. On a~given computational basis, it creates a superposition state. This operation can be illustrated as a rotation \(\frac{\widehat x + \widehat z}{\sqrt{2}}\) on the Bloch sphere by an angle \(\pi\). This gate is given by a unitary matrix:

\begin{equation}
\label{eqn:diracstate_4}
H = \frac{1}{\sqrt2}\begin{bmatrix}
1 & 1\\
1 & -1
\end{bmatrix}
\end{equation}

Phase gates also fall into the category of single-cubit gates. U1 gate rotates the cubit around \(z\) axis, U2 gate around \(x+z\) axis and U3 gate is a rotation gate with three Euler angles. The matrix representations of these gates are as follows:

\begin{equation}
\label{eqn:diracstate_5}
U1(\theta) = \frac{1}{\sqrt2}\begin{bmatrix}
1 & 0\\
0 & e^{i\frac{\theta}{2}}
\end{bmatrix}
\end{equation}

\begin{equation}
\label{eqn:diracstate_6}
U2(\theta, \gamma) = \begin{bmatrix}
1 & -e^{i\frac{\gamma}{2}}\\
e^{i\frac{\theta}{2}} & e^{i\frac{\theta+\gamma}{2}}
\end{bmatrix}
\end{equation}

\begin{equation}
\label{eqn:diracstate_7}
U3(\lambda, \theta, \gamma) = \begin{bmatrix}
\cos\frac{\lambda}{4} & -e^{i\frac{\gamma}{2}}\sin\frac{\lambda}{4}\\
e^{i\frac{\theta}{2}}\sin\frac{\lambda}{4} & e^{i\frac{\theta+\gamma}{2}}\cos\frac{\lambda}{4}
\end{bmatrix}
\end{equation}

For two-qubit gates consisting of a control qubit and a target qubit, the state of the control qubit remains constant during operations. 
A change of the state of the target qubit is performed when the state of the control qubit is \(|1 \rangle\). 
A popular one in this group is the \(CNOT\) gate (controlled Pauli-X). 
The unitary matrix representing the CNOT gate is of the form:

\begin{equation}
\label{eqn:diracstate_8}
CNOT = \begin{bmatrix}
1 & 0 & 0 & 0\\
0 & 1 & 0 & 0\\
0 & 0 & 0 & 1\\
0 & 0 & 1 & 0
\end{bmatrix}
\end{equation}

For the CZ gate, the representation is as follows:
\begin{equation}
\label{eqn:diracstate_9}
CZ = \begin{bmatrix}
1 & 0 & 0 & 0\\
0 & 1 & 0 & 0\\
0 & 0 & 1 & 0\\
0 & 0 & 0 & -1
\end{bmatrix}
\end{equation}

In the case of a controllable CRY rotation gate, the matrix is given:
\begin{equation}
\label{eqn:diracstate_10}
CRY(\theta) = \begin{bmatrix}
1 & 0 & 0 & 0\\
0 & \cos\frac{\theta}{2} & 0 & -\sin\frac{\theta}{2}\\
0 & 0 & 1 & 0\\
0 & \sin\frac{\theta}{2} & 0 & \cos\frac{\theta}{2}
\end{bmatrix}
\end{equation}

A network built solely from quantum layers based on unitary gates was first proposed by Edward Farhi and Hartmut Neven in 2018 \cite{farhi}. 
In this neural network, the quantum circuit consisted of a sequence of parameter-dependent unary transformations that performed transformations on the input quantum state.
At the current stage of development, hybrid networks seem much more promising because they assume operations on small quantum circuits with negligible or zero error correction, so they are shown to have potential for use with upcoming quantum computer architectures. This is because Noisy Intermediate-Scale Quantum (NISQ) era computer architectures are solutions with limited computational resources and need the support of performing some of the computation on classical hardware. 
In many publications, the term \emph{quantum neural network} is actually a hybrid model. 
It is worth noting that, at the current stage of development, purely quantum architectures are~successfully applied only to classification problems with a small number of classes, since the number of qubits required to encode and read data increases with the number of classes. 
For some classification tasks, this leads to the system exceeding the number of available qubits or generating significant errors.

\section{Previous work}
\label{previouswork}


Due to their fast convergence and high accuracy, quantum neural networks can find applications in many areas, especially in vision systems, whose bottleneck is the need to process large amounts of data efficiently. 
For example, in the paper \cite{boyda} a successful attempt was made to recognize trees in the aerial space of California, where ensemble methods were used. 
The authors proposed a truncation and rescaling of the training objective through a trainable metaparameter. 
Accuracies of 92\(\%\) in validation and 90\(\%\) on a test scene were obtained. \par

In the paper \cite{li} experiments on cancer recognition were carried out. 
The authors proposed a qubit neural network model with sequence input based on controlled rotation gates, called QNNSI. 
The three-layer model with a hidden layer, which employed the Levenberg-Marquardt algorithm for learning, was the best of all the approaches tested. 
The experimental results reveal that a greater difference between the input nodes and the sequence length leads to a lower performance of the proposed model than that of the classical ANN, in contrast, it obviously enhances the approximation and generalization ability of the proposed model when the input nodes are closer to the sequence length \cite{li}. \par

An experiment proposed in \cite{yu} included the classification of vehicles.  
Successful object segmentation proposed in \cite{aytekin}, \cite{aytekin2}. 
Other research shows that quantum neural networks can be effective in the recognition of pneumonia \cite{yumin}.
The classification of ants and bees introduced in \cite{mari} indicates that the use of a transfer learning technique can result in the creation of an efficient system for differentiating between two types of insects.
The retinal color fundus images and chest radiography classification presented in \cite{mathur} is another example of the superiority of quantum neural networks over classical ones.

Topics similar to the subject of this research can be found in the paper \cite{majumder}. 
The authors applied a transfer learning approach and used a Resnet18 network trained on an ImageNet set, replacing the fully connected layer with a quantum layer and a PyTorch layer-enabled module. 
The results obtained from such a~network were compared with those of a simple two-layer classical network. 
Using the PennyLane plug-in, quantum computations were possible, and the popular PyTorch library was used. 
Three steps are distinguished in the quantum layer. 
The first was to embed the data in a quantum circuit using a single or combinations of several gates: Hadamard gate, Rotational Y, Rotational X,
Rotational Z, U1, U2, and U3 gates. 
The data were then fed into a parameterized quantum circuit of two-qubit gates (Controlled NOT, Controlled Z, and Controlled RX) with single-qubit parameterized gates (i.e., Rotational Y, Rotational X, Rotational Z, U1, U2 and U3). 
The last was a measurement step using bases composed of X, Y, Z quantum gates. 
In this experiment, the LISA traffic sign set was reduced to 18 traffic sign classes. 
A binary classification based on the network's recognition of stop signs and the class in which the other signs were placed was tested. 
This used 231 samples divided into a training set and a test set in a ratio of 80:20.
For multiclass classification, the signs were assigned to one of the three classes: stop signs, speed limit signs, and the class containing the other signs. 
To do this, 279 samples divided into training and test set were used in a ratio of 80:20. 
The following attack algorithms were used: the gradient attack, Fast Gradient Sign Method (FGSM), Projected Gradient Descent (PGD) attack. 
The implementations used achieved high accuracy (more than 90\%).

\section{QNN for traffic sign recognition}
\label{qnnfortsr}


The GTSRB (German Traffic Sign Recognition Benchmark) dataset used in this experiment was created in 2011. It contains 39209 training images and 12630 test images, which varies in size and other parameters, e.g. brightness and contrast. Each of the images belongs to one of the 43 class representing traffic signs. \par

\begin{figure}[!t]
\includegraphics[scale=0.5]{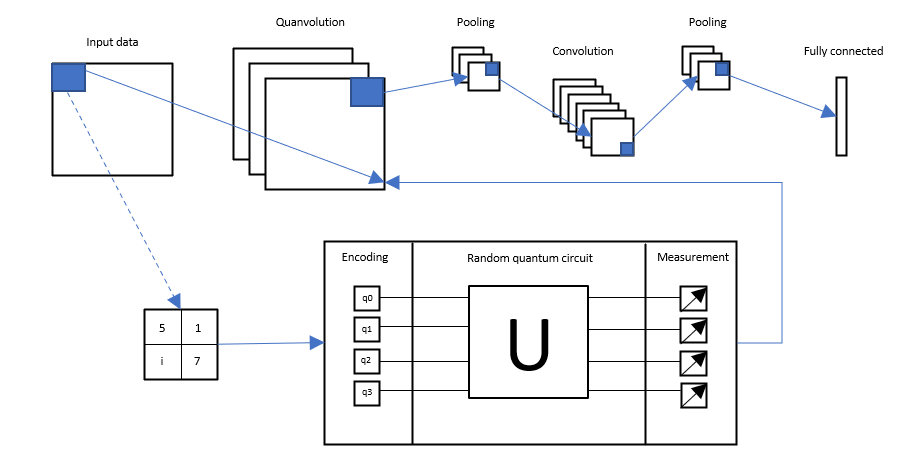}
\centering
\caption{The architecture of a quanvolutional neural network proposed in \cite{henderson}. The quanvolution layers are made up of a group of \(N\) quantum filters that create feature maps. Local subareas of data are processed spatially using random or structured quantum circuits.
\label{fig:hybrid}}
\end{figure}

An initial pre-processing consisting of two stages was introduced. 
In the first, grayscale normalization was performed and in the second, the dataset was filtered and only images larger than \(64\times64\) pixels were considered in the next steps.
This was due to the observation that the network performed better for larger images. 
In the next step, the data set was divided into the 80:10:10 ratio for the training, test and validation datasets.
Data labels were converted to vectors of length equal to the number of classes present according to the one-hot encoding method. 
The images of each subset were then subjected to a quantum convolution operation. 
In~experiments, a similar approach to that presented in \cite{henderson} was used. 
Quantum convolution (quanvolution) layers were built from a~group of \(N\) quantum filters that create feature maps by locally transforming the input data. 
The key difference between quantum and classical convolution is the processing of spatially local subareas of data using random or structured quantum circuits. 
Figure \ref{fig:hybrid} shows the design of such an architecture.

For each image, a small area of~dimensions \(2\times2\) was embedded in the quantum circuit by rotations parameterized by the rotation angle of the qubits initialized in the ground state. 
Further quantum computations were performed on the basis of the unitary space generated by a random circuit similar to that proposed in \cite{henderson}.
The experiment performed in~this paper uses a circuit consisting of a single-qubit rotation operator \(Y\) and \emph{Pauli-X} gate. 
The number of random layers was set to two because this parameter was not observed to lead to any change in classification quality during the experiments. 
Then, as in classical convolution, each z~value was mapped to each of the output z~channels. 
Iteratively repeating these steps yielded a complete output object that could be structured into a multichannel image. 
The main difference between the experiment proposed in this paper versus \cite{henderson} was to reduce the number of epochs from ten thousand to one hundred, reduce the dataset to seven thousand samples instead of dimensionality reduction, and change the neural network architecture to a~more advanced one. 

Both classical and quantum networks consisted of convolutional, max-pooling, dropout, flatten, and dense layers. The difference between them was in the convolution operation performed on the images. In the hybrid model, images convolved with quantum circuits were fed. 
The models were compiled with the adaptive optimizer Adam and the cost function as the cross-entropy with the accuracy metric. 
The final step was to verify the quality of the learning by determining the confusion matrix and the precision, recall, and f-beta parameters on the test set. 
Table \ref{tab:results} contains the results obtained.

\begin{table}[!t]
\caption{Summary of results obtained for the classical network and the network with quantum convolution at particular batch sizes for the 43 classes of the GTSRB dataset}
\label{tab:results}
\begin{adjustbox}{width=1\textwidth}
\small
\begin{tabular}{| c | c | c | c | c | c | c | c | c |}
\hline
\multicolumn{1}{|c}{\textbf{Batch}} & \multicolumn{2}{|c}{\textbf{Accuracy}} & \multicolumn{2}{|c}{\textbf{Precision}} & \multicolumn{2}{|c|}{\textbf{Recall}} & \multicolumn{2}{c|}{\textbf{F-beta}} \\

\textbf{size} & CNN & QNN & CNN & QNN & CNN & QNN & CNN & QNN  \\
\hline
4 &  0.9957 & 0.9254 & 0.9914 & 0.9211 & 0.9918 & 0.9310 & 0.9912 & 0.9216 \\
\hline
8 &  0.9914 & 0.9426 & 0.9949 & 0.9426 & 0.9943 & 0.9456 & 0.9944 & 0.9216 \\
\hline
16 &  0.9971 & 0.9426 & \textbf{1.0000} & 0.9426 & \textbf{1.0000} & 0.9453 & \textbf{1.0000} & 0.9419 \\
\hline
32 &  \textbf{0.9986} & 0.9369 & 0.9947 & 0.9947 & 0.9369 & \textbf{0.9943} & 0.9476 & 0.9403 \\
\hline
64 &  0.9957 & 0.9383 & 0.9959 & \textbf{0.9983} & 0.9957 & 0.9423 & 0.9954 & 0.9378 \\
\hline
128 &  \textbf{0.9986} & \textbf{0.9440 }& 0.9973 & 0.9440 & 0.9986 & 0.9494 & 0.9979 & \textbf{0.9446} \\
\hline
256 & 0.9943 & 0.9369 & 0.9973 & 0.9369 & 0.9971 & 0.9428 & 0.9971 & 0.9382 \\
\hline
512 &  \textbf{0.9986} & 0.9354 & 0.9959 & 0.9354 & 0.9957 & 0.9381 & 0.9956 & 0.9351 \\
\hline
\end{tabular}
\end{adjustbox}
\end{table}

In~the case of the classical network, there was only one incorrect classification for a given input, and the number of images assigned to incorrect labels did not exceed one. 
For the network with quantum convolution, the incorrect detections are slightly more, and those that recur most frequently and contain more than one misclassified label for a given batch are the following pairs of visually similar characters:

\begin{itemize}
    \item pedestrians and priority road,
    \item bicycles crossing and double curve,
    \item wild animals and road narrows on the right,
    \item go straight or right and pedestrians,
    \item go straight or left and pedestrians,
    \item wild animals and pedestrians,
    \item road narrows right and end of all limits,
    \item pedestrians and road narrows right.
\end{itemize}

In some of the above cases, there is little or no similarity from the point of view of the human viewer, for example ``crossing priority at an intersection'' and ``attention pedestrians''. The implication is that the process of quantum convolution makes the model more robust to errors associated with similar sign content, while shifting errors towards a less recognized cause. This brings back an association with characteristic quantum non-explicability.

The advantage of the classical network can be seen each time, as it achieves up to 99.86\(\%\) accuracy for batch sizes of 32, 128, 512. For the network with quantum convolution, the highest obtained result for accuracy is 94.40\(\%\) for batch size 128, so for the best cases this network shows almost 4\(\%\) worse classification accuracy. Although the classical network in the worst case achieved an accuracy of 99.14\(\%\) for a batch of size 8, the network with quantum convolution achieves a decrease greater than 2\(\%\) in classification quality. 

\begin{figure}[t!]
\includegraphics[scale=0.3]{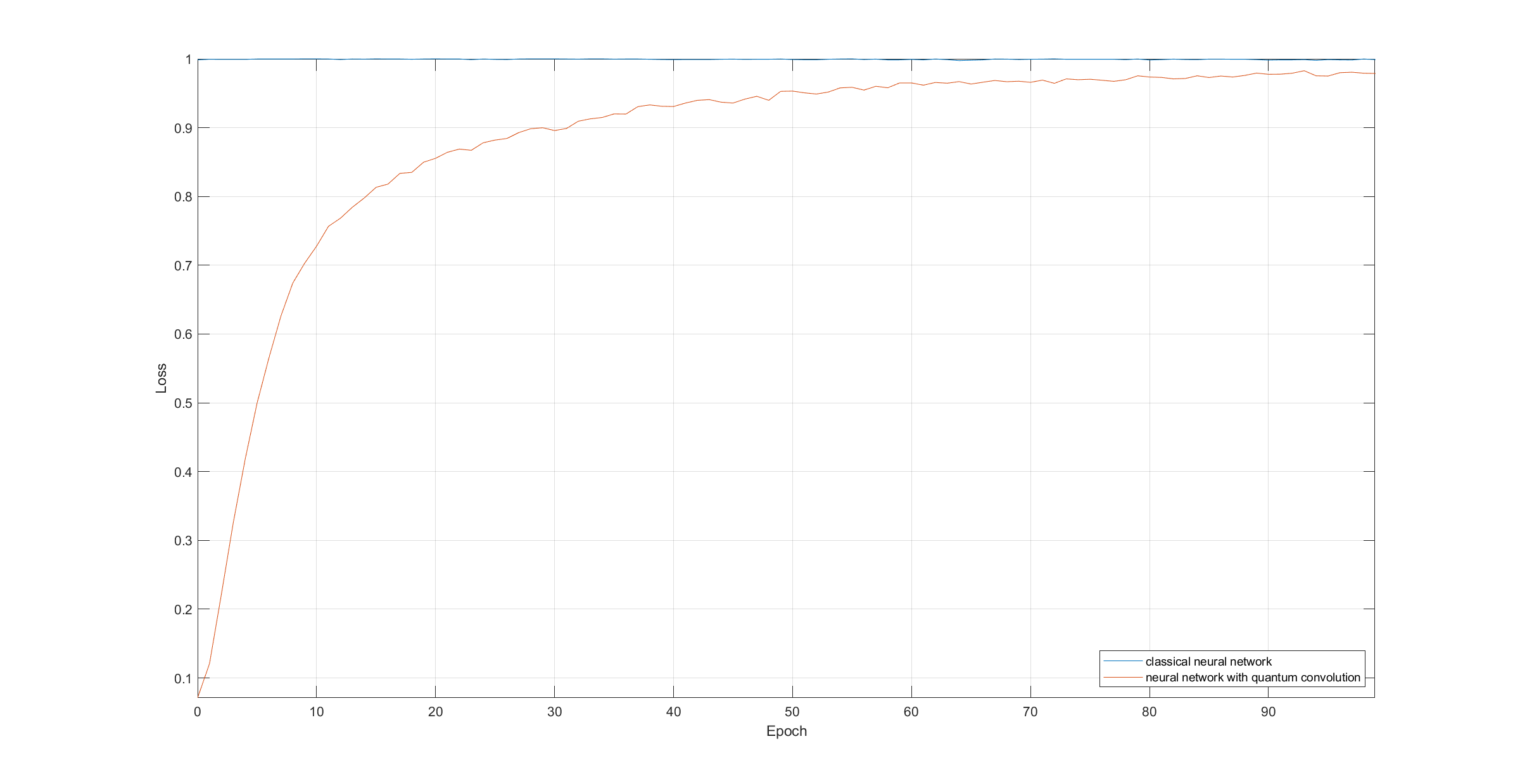}
\centering
\caption{Comparison of the accuracy obtained with a batch size of 128 for the classical network and the network with quantum convolution on the training set}
\label{fig:gtsrb_comparison_quanvolution_classical_accuracy_over_batch128}
\end{figure}

\begin{figure}[t!]
\includegraphics[scale=0.4]{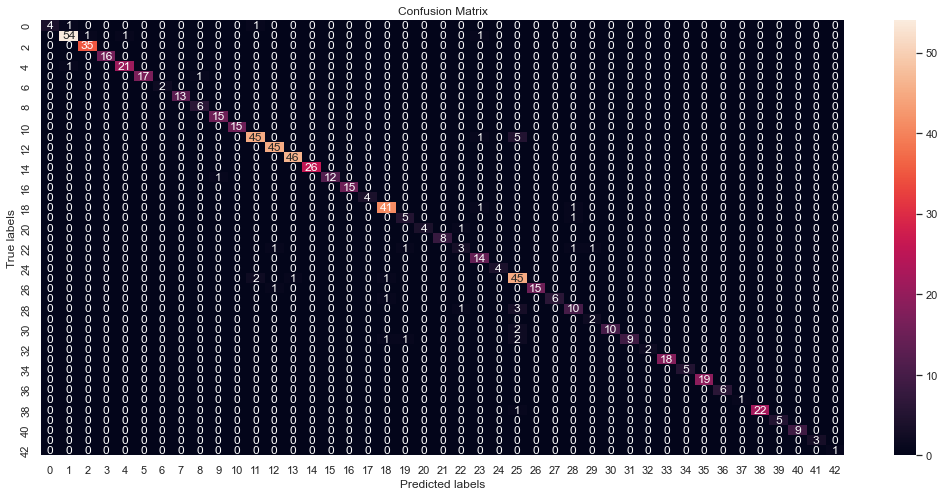}
\centering
\caption{Confusion matrix with a batch of size 128 for networks with quantum convolution}
\label{fig:quanvolutional_network_confusion_matrix_epochs_100_batch_size_128}
\end{figure}

In~the case of the network with quantum convolution, this is 94.4\(\%\) for a batch size of 128, which is more than 5\% lower. The learning flow for this batch is illustrated in Fig. \ref{fig:gtsrb_comparison_quanvolution_classical_accuracy_over_batch128}. The classical network achieves a~worst-case score of 99.14\(\%\) for a~batch of size 4, and for the network with~quantum convolution, this score for the same batch is more than 2\(\%\) worse. The differences in the values of the recall and f-beta coefficients were at a similar level. Figure \ref{fig:quanvolutional_network_confusion_matrix_epochs_100_batch_size_128} shows the confusion matrix for a neural network with quantum convolution trained with a batch of size 128.

\begin{figure}[t!]
\includegraphics[scale=0.3]{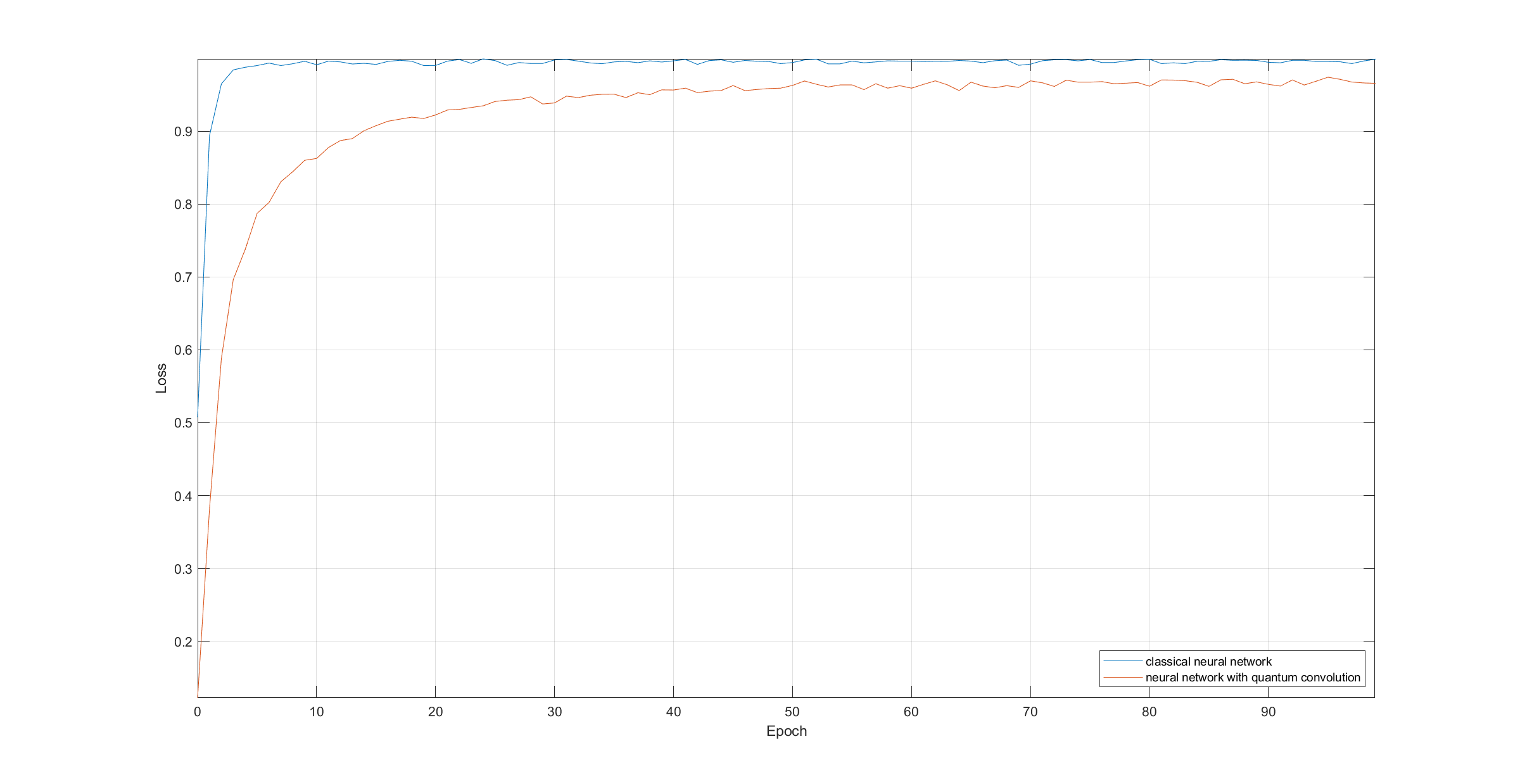}
\centering
\caption{Comparison of the accuracy obtained with a batch size of 16 for the classical network and the network with quantum convolution on the training set}
\label{fig:gtsrb_comparison_quanvolution_classical_accuracy_over_batch16}
\end{figure}

The analysis shows that for both networks the worst results were obtained with a batch size equal to 4. 
The best results for each network were obtained with different batch sizes. 
For the classical network, the best of the tested ones was a batch size of 16, while for the network with quantum convolution it was a batch size of 128. 
This results in an important difference in the approach to quantum networks, as these models can take larger portions of data, and thus their learning will be faster. 
Note, however, that they are very sensitive to the appropriate choice of batch file, much more so than classical networks, which show degradation at the tenths of a percent level over their best and worst performance.

The algorithm applied allowed us to obtain much better classification accuracy than it was in the case of \cite{potempa}, whereby in our experiment, classification was performed using 43 classes. The results obtained are at a level similar to some of the experiments carried out in \cite{majumder}.

\section{Conclusion}
\label{conclusion}


Quantum machine learning is a challenging field. 
This concerns not only the proper design of algorithms but also the very way in which classical real data are encoded in quantum space. 
Uncertainty is also provided by the fact that the only official form of a quantum computer is not known, but rather a number of proposals for what it could look like. 
This results in potentially going down wrong paths and having to redesign the algorithms multiple times. 

The experiment showed that it is possible to achieve a high classification accuracy (more than 94\%) for a neural network with quantum convolution, but raised the question of quantum supremacy. Quantum algorithms require special preprocessing on the dataset, which in the case of hybrid networks, of which the network with quantum convolution is an example, requires significant computational resources, and the results obtained do not show the expected significant superiority over the results returned by the classical network. 
However, the purpose of the experiment was not to improve the chosen path, but to outline a direction on how to prototype an exemplary quantum-classical neural network model for the multiclass classification problem, which could find its application in vision systems.

The prospects for further development of the project are very broad. 
It is planned to try to compare the influence of data augmentation on learning accuracy and the potential overfitting effect in the comparison of the classical network and the network with quantum convolution. In addition, another task will be undertaken in the~area of explainable AI, that is, an attempt to visualize, for example, in the form of a heat map, which features of the image influence a~given classification result. 
It would also be interesting to try out the designed network architecture on a real quantum computer along with undertaking an analysis of the temporal performance of the algorithm. 
\end{sloppypar}

\subsubsection{Acknowledgements} The work presented in this paper was supported by the AGH University of Science and Technology project no. 16.16.120.773.



%
%
%
%





\bibliographystyle{splncs03_fixed}
\bibliography{SK_TK_ICCVG2022}

\end{document}